\documentclass{article}

\usepackage[preprint, nonatbib]{neurips_data_2023}

\usepackage[utf8]{inputenc} 
\usepackage[T1]{fontenc}    
\usepackage{hyperref}       
\usepackage{url}            
\usepackage{booktabs}       
\usepackage{amsfonts}       
\usepackage{nicefrac}       
\usepackage{microtype}      
\usepackage{xcolor}         
\usepackage{threeparttable}
\usepackage{color}
\usepackage{makecell}
\usepackage{multirow}
\usepackage{subfigure}
\usepackage{graphicx}
\usepackage{wrapfig}
\usepackage{algorithmic}
\usepackage{algorithm}
\usepackage{enumitem}

\newtheorem{Def}{Definition}

\title{arXiv4TGC: Large-Scale Datasets for Temporal Graph Clustering}

\author{%
Meng Liu \quad Ke Liang \quad Yue Liu \quad Siwei Wang \quad Sihang Zhou \quad Xinwang Liu\thanks{Corresponding author.} \\
National University of Defense Technology\\
Changsha, China\\
\texttt{mengliuedu@163.com, xinwangliu@nudt.edu.cn} \\
}

\begin{document}

\maketitle

\begin{abstract}
Temporal graph clustering (TGC) is a crucial task in temporal graph learning. Its focus is on node clustering on temporal graphs, and it offers greater flexibility for large-scale graph structures due to the mechanism of temporal graph methods. However, the development of TGC is currently constrained by a significant problem: the lack of suitable and reliable large-scale temporal graph datasets to evaluate clustering performance. In other words, most existing temporal graph datasets are in small sizes, and even large-scale datasets contain only a limited number of available node labels. It makes evaluating models for large-scale temporal graph clustering challenging. To address this challenge, we build arXiv4TGC, a set of novel academic datasets (including arXivAI, arXivCS, arXivMath, arXivPhy, and arXivLarge) for large-scale temporal graph clustering. In particular, the largest dataset, arXivLarge, contains 1.3 million labeled available nodes and 10 million temporal edges. We further compare the clustering performance with typical temporal graph learning models on both previous classic temporal graph datasets and the new datasets proposed in this paper. The clustering performance on arXiv4TGC can be more apparent for evaluating different models, resulting in higher clustering confidence and more suitable for large-scale temporal graph clustering. The arXiv4TGC datasets are publicly available at: \url{https://github.com/MGitHubL/arXiv4TGC}.
\end{abstract}

\section{Introduction}

Temporal graph clustering is a vital component of graph clustering, which involves clustering nodes in temporal graph datasets \cite{cui2018survey, WLH2020GRL}. Existing graph clustering methods focus mainly on static graphs \cite{yue2022survey, liuyue_DCRN} and ignore the temporal information between nodes, which is prevalent in real-world data \cite{trivedi2019dyrep, liang2022reasoning, wang2021inductive}. It means that in many graph clustering related applications, temporal graphs will be trained as static graphs, resulting in the loss of dynamic information and significant computational consumption \cite{gao2021equivalence, wu2021sagedy, pareja2020evolvegcn}. 

Recently, the TGC framework \cite{TGC_ML}, which comprehensively focuses on temporal graph clustering, identified the major challenge hindering the development of temporal graph clustering: a lack of publicly temporal graph datasets available for node clustering. Although temporal graph clustering can be processed under the unsupervised scenario, node labels are necessary to evaluate the experimental performance. Unfortunately, the vast majority of temporal graph datasets suffer from one of the following problems:

\begin{itemize}[leftmargin=0.7cm]
\item[(1)] Many temporal graph models are designed under the task of link prediction task that does not require node labels, because many datasets with no labels or only two labels (lots of 0 and a few 1). As a result, models on these datasets require no label information (i.e., without considering node clustering) or are designed to predict whether some nodes are active at some given moments. These datasets with only two types of labels are more suitable to be used for binary classification compared to multi classification, making them unavailable for node clustering, e.g., CollegeMsg, LastFM, Wikipedia, Reddit, Ubuntu, and MOOC.

\item[(2)] Node labels in some datasets do not match the nodes' characteristics. To name a few, different product ratings by users can be considered as labels (usually from 1 to 5), but it can hardly claim that these rate values are more relevant to the product characteristics than the product category labels, resulting in low clustering performance on these datasets, e.g., Bitcoin, MovieLens-1M, Yelp, Tmall, and Amazon.

\item[(3)] Only a small number of datasets are suitable for temporal graph clustering, and these are small in size, e.g., DBLP, Brain, and Patent. (We will later provide a detailed comparison and discussion of these datasets.)
\end{itemize}

To address the aforementioned challenges, we have developed a set of large-scale academic datasets, arXiv4TGC, to facilitate the investigation of temporal graph clustering. This dataset comprises five sub-datasets, namely arXivAI, arXivCS, arXivMath, arXivPhy, and arXivLarge, which are extracted from the arXiv open platform and cover 172 subfields\footnote{\url{https://arxiv.org/category_taxonomy}} classified by arXiv. We consider papers as nodes and the citations between papers as edges. When a paper cites another paper at some point in time, we regard an interaction between the two nodes to have occurred. The largest sub-dataset, arXivLarge, contains 1.3 million labeled nodes and 13 million temporal edges. The information in these datasets is anonymized and easily accessible to researchers. In summary, our contributions include:

\begin{itemize}[leftmargin=0.7cm]
\item We point out the absence of existing benchmarks for large-scale temporal graph clustering. Motivated by it, we build arXiv4TGC, a set of novel academic datasets for large-scale temporal graph clustering.

\item We describe the details of arXiv4TGC, which contain five datasets, \textit{i.e.,} arXivAI, arXivCS, arXivMath, arXivPhy, and arXivLarge, in different scale requirements. Note that the arXiv4TGC datasets are publicly available at: \url{https://github.com/MGitHubL/arXiv4TGC}.

\item We further compare the clustering performance with typical temporal graph learning models on both previous public datasets and the new datasets proposed in this paper. The clustering performance on arXiv4TGC can be more apparent for evaluating different models, resulting in higher clustering confidence and more suitable for large-scale temporal graph clustering.
\end{itemize}

The rest of this paper is organized as follows. In Section 2, we give a brief definition of temporal graph clustering. In Section 3, we compare different temporal graph datasets. In Section 4, we describe the details of arXiv4TGC. In Section 5, we conclude this work and discuss future work.

\section{Temporal graph clustering}

Temporal graph clustering, also known as node clustering on temporal graphs, aims to group nodes into different clusters. If a graph records node interactions with time information, we denote it as a temporal graph  \cite{nguyen2018continuous, xu2020inductive, MNCI_ML_SIGIR}.

\begin{Def}
\textbf{Temporal graph.}
Given a temporal graph $G=(V, E, T)$, where $V$ is the set of nodes and $E$ is the set of node interactions. Note that the concept of edge in static graphs is replaced by interaction here. This is because multiple interactions may occur at different timestamps on an edge between two nodes. A temporal graph is represented as an interaction sequence (adjacency list), which consists of a large number of $(node, node, time)$ ordered by time.

Temporal graph clustering aims to divide the nodes in the temporal graph into $K$ groups. This is a two-step process: The temporal is first modeled to generate the node representations, then these representations are fed into the K-means model for clustering and evaluation.
\end{Def}

Although there are many methods that focus on temporal graphs, most of them usually evaluate their performance on link prediction tasks than node clustering tasks. One of the major challenges hindering the development of temporal graph clustering is that only a small number of datasets are suitable for node clustering. Next, we will discuss in detail the existing sequential graph datasets and explain why most of them are not suitable for temporal graph clustering.

\section{Related datasets}

\begin{wrapfigure}{r}{0.55\textwidth}
\includegraphics[width=0.55\textwidth]{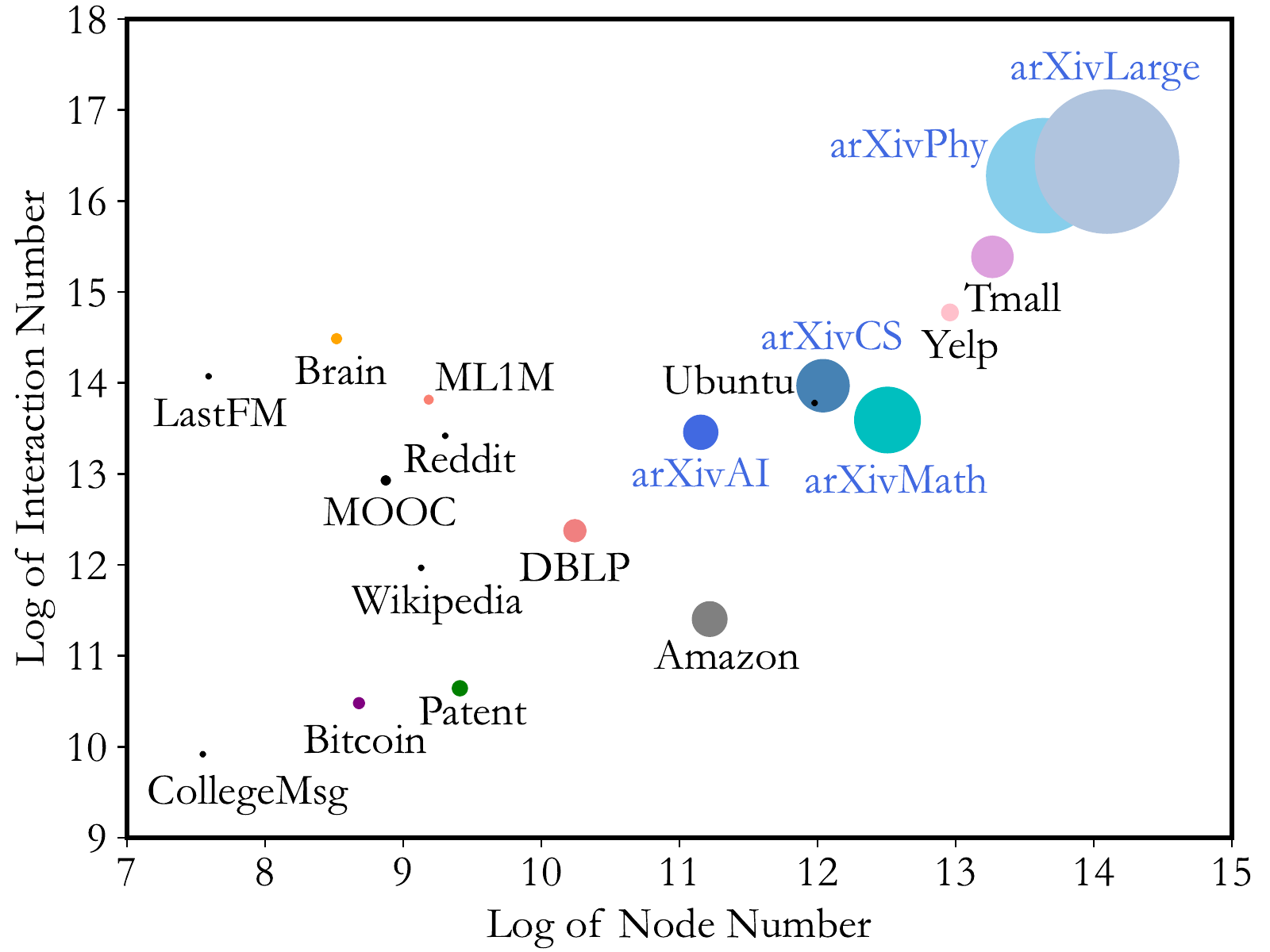}
\caption{Scatter plots for different dataset sizes. The horizontal axis represents the node number and the vertical axis represents the interaction number. Since the variation between different data sets is too large, we present the data size in logarithmic form. The size of the scatter represents the number of node labels available for the dataset.}
\label{scatter}
\end{wrapfigure}

In this section, we make a comparison of multiple temporal graph datasets that are publicly available. Fig. \ref{scatter} displays the varying dataset sizes. It is noteworthy that the dataset sizes have been logarithmized. This has been done to ensure that datasets are easily distinguishable. If the original data magnitude is used to represent these datasets, most of the datasets will be concentrated in the lower left quadrant and it will become difficult to differentiate them. The exact sizes of these datasets can be obtained from Table \ref{datasets}.

These datasets are all publicly available datasets: CollegeMsg \cite{panzarasa2009patterns} comprises private messages exchanged on an online social network at the University of California, Irvine. LastFM \cite{kumar2019predicting} contains one month's worth of information on who listens to which songs. Wikipedia \cite{kumar2019predicting} is a collection of edits made by users on Wikipedia pages over a month. Reddit \cite{kumar2019predicting} includes one month's worth of posts made by users on subreddits. Ubuntu \cite{paranjape2017motifs} is a temporal graph of questions and answers on the Ask Ubuntu website of the Stack Exchange platform. MOOC \cite{kumar2019predicting} consists of actions taken by students while participating in a MOOC online course, such as viewing a video or submitting an answer. Bitcoin \cite{kumar2016edge} is bitcoin trading graph on the Bitcoin OTC platform, and it represents who trusts whom. MovieLens-1M (ML1M) \cite{li2020time} is a widely used dataset containing movie ratings. Amazon \cite{ni2019justifying} is a magazine subscription graph dataset from the Amazon platform. Yelp \cite{zuo2018embedding} is a user rating graph dataset from the Yelp platform. Tmall \cite{zuo2018embedding} comes from the Tmall platform, which extracts the user product purchase data during the "Double-Eleven" shopping campaign. Brain \cite{preti2017dynamic} is a connectivity graph of brain tissue in humans. Patent \cite{hall2001nber} is a graph of patent citations. DBLP \cite{zuo2018embedding} is a dataset of co-authorship in the Computer Science domain taken from the DBLP website.

According to Fig. \ref{scatter} and Table \ref{datasets}, we can divide these publicly available datasets into three parts: \textbf{unlabeled} datasets (CollegeMsg, LastFM, Wikipedia, Reddit, Ubuntu, and MOOC), \textbf{labeled untrustworthy} datasets (Bitcoin, ML1M, Amazon, Yelp, and Tmall), and \textbf{small-scale labeled} datasets (Brain, Patent, DBLP). Most of these datasets are small-scale, a few larger datasets belong to the case unlabeled (Ubuntu) or labeled untrustworthy (Amazon, Yelp, and Tmall). This means that there is no very suitable dataset for the large-scale temporal graph clustering task. To further compare these datasets, we report multiple experimental results in Section \ref{Exp}.

\begin{table}[t]
\fontsize{10}{12.5}\selectfont 
\caption{Comparison of publicly available temporal graph datasets.}
\centering
\begin{tabular}{c|ccccc}
    \toprule[2pt]
    Dataset    & Nodes     & Interactions & Class & Labels & Timestamps   \\
    \midrule[1pt]
    CollegeMsg   & 1,899     & 20,296       & N/A     & N/A & 193    \\
    LastFM   & 1,980     & 1,293,103    & N/A     & N/A &  30      \\
    Wikipedia      & 9,228     & 157,474      & N/A     & N/A & 30  \\
    Reddit     & 10,985    & 672,447      & N/A     & N/A  &  30    \\
    Ubuntu     & 159,316   & 964,437      & N/A     & N/A &  2,613      \\
    MOOC       & 7,144     & 411,749      & N/A     & N/A &  -      \\
    \midrule[1pt]
    Bitcoin    & 5,881     & 35,592       & 21    & 5,858 &  27,487  \\
    ML1M       & 9,746     & 1,000,209    & 5     & 3,706  &  25,212 \\
    Amazon     & 74,526    & 89,689       & 5     & 72,098 & 5,804  \\
    Yelp     & 424,450   & 2,610,143    & 5     & 15,154 & 153  \\
    Tmall     & 577,314   & 4,807,545    & 10    & 104,410 & 186 \\
    \midrule[1pt]
    Brain      & 5,000     & 1,955,488    & 10    & 5,000  &  12 \\
    Patent    & 12,214    & 41,916       & 6     & 12,214 &  891 \\
    DBLP      & 28,085    & 236,894      & 10    & 28,085 & 27  \\
    \midrule[1pt]
    arXivAI    & 69,854    & 699,206      & 5     & 69,854  & 27 \\
    arXivCS    & 169,343   & 1,166,243    & 40    & 169,343 & 29 \\
    arXivMath  & 270,013   & 799,745      & 31    & 270,013 & 31 \\
    arXivPhy   & 837,212   & 11,733,619   & 53    & 837,212 & 41 \\
    arXivLarge & 1,324,064 & 13,701,428   & 172   & 1,324,064 & 41 \\
    \toprule[2pt]
\end{tabular}
\label{datasets}
\end{table}

\section{arXiv4TGC datasets}

In this section, we describe our arXiv4TGC datasets in detail. We first introduce the data processing flow, and then the details of each dataset. Since the experimental results are given above, they are not listed here.

\subsection{Data processing}
\label{processing}

Our arXiv4TGC datasets include 5 temporal graph datasets: arXivAI, arXivCS, arXivMath, arXivPhy, and arXivLarge. In these datasets, we denote papers as nodes, and citations between papers as interactions, where the timestamps of interactions depend on the citation time of the paper.

The source data of arXiv4TGC was obtained from the publicly available dataset ogbn-papers100M, which was curated by OGB \cite{wang2020microsoft}. However, this dataset contains a significant amount of unlabeled data and is too extensive to be used conveniently for simple testing. Therefore, we performed additional processing on top of ogbn-papers100M to extract several academic datasets that are appropriate for temporal graph clustering:

\begin{itemize}[leftmargin=0.7cm]
\item[(1)] We initially extracted the node interaction information from the original data, identified their corresponding node labels, and removed the unlabeled nodes along with the corresponding interactions (arXivLarge).

\item[(2)] Next, we extracted some edges into a new dataset based on the domain to which the node labels belonged. During this process, we considered different domain combinations and ultimately chose to retain four combinations (arXivAI, arXivCS, arXivMath, and arXivPhy).

\item[(3)] Since the node numbers in the extracted datasets were scattered, we renumbered them and updated the node label list.

\item[(4)] Additionally, some temporal graph methods do not consider node characteristics. To ensure a fair comparison, we provided node features based on position encoding \cite{vaswani2017attention} in addition to the original features.

\item[(5)] Finally, we consolidated and compressed these datasets for public download. We also provided the HTNE model that was applied to the temporal graph clustering task and made it easy to utilize the arXiv4TGC datasets to obtain benchmark results.

\end{itemize}

\subsection{Data details}

\begin{table}[t]
\fontsize{9.5}{14}\selectfont 
\caption{Research subfields included in the different datasets.}
\centering
\resizebox{\linewidth}{!}{
    \begin{tabular}{c c}
        \toprule[2pt]
        Dataset      & Research Subfields    \\
        \midrule[1pt]
        arXivAI         & \makecell[c]{Artificial Intelligence (\textbf{arxiv cs ai}), Machine Learning (\textbf{arxiv cs lg}), Computation and Language (\textbf{arxiv cs cl}),\\ Computer Vision and Pattern Recognition (\textbf{arxiv cs cv}), Neural and Evolutionary Computing (\textbf{arxiv cs ne})} \\  
        \midrule[1pt]
        arXivCS & \makecell[c]{\textbf{arxiv cs:} na, mm, lo, cy, cr, dc, hc, ce, ni, cc, ai, ma, gl, ne, sc, ar, cv, gr,\\ et, sy, cg, oh, pl, se, lg, sd, si, ro, it, pf, cl, ir, ms, fl, ds, os, gt, db, dl, dm} \\
        \midrule[1pt]
        arXivMath & \makecell[c]{\textbf{arxiv math:} ac, ag, ap, at, ca, co, ct, cv, dg, ds, fa, gm, gn, gr,\\ gt, ho, kt, lo, mg, na, nt, oa, oc, ph, pr, qa, ra, rt, sg, sp, st}\\
        \midrule[1pt]
        arXivPhy & \makecell[c]{\textbf{arxiv:} gr qc, hep ex, hep lat, hep ph, hep th, math ph, nucl ex, nucl th, quant ph, \\ \textbf{arxiv astro ph:} \emph{self}, co, ep, ga, he, im, sr, \, \textbf{arxiv nlin}: ao, cd, cg, ps, si, \\ \textbf{arxiv cond mat:} \emph{self}, dis nn, mes hall, mtrl sci, other, quant gas, soft, stat mech, str el, supr con, \\ \textbf{arxiv physics}: acc ph, ao ph, app ph, atm clus, atom ph, bio ph, chem ph, class ph, comp ph, data an, \\ ed ph, flu dyn, gen ph, geo ph, hist ph, ins det, med ph, optics, plasm ph, pop ph, soc ph, space ph}\\
        \midrule[1pt]
        arXivLarge & \makecell[c]{All 172 research areas in the fields of Computer Science, Physics, Economics Science, Mathematics,\\ Electrical Engineering and Systems, Quantitative Biology, Quantitative Finance, and Statistics.}\\
        \bottomrule[2pt]     
\end{tabular}}
\label{fields}
\end{table}

\begin{figure}[t]
\centering
\subfigure[Class Distribution on arXivAI]{
    \includegraphics[width=0.32\textwidth]{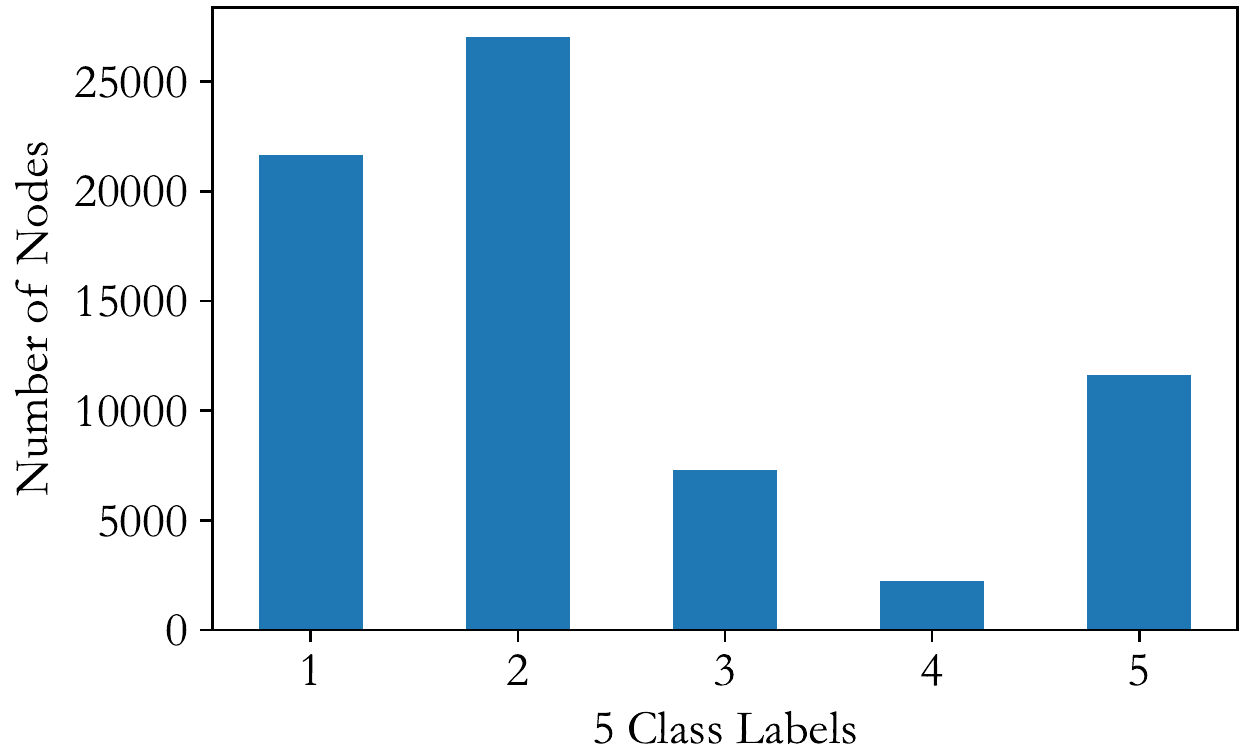}}
\subfigure[Class Distribution on arXivCS]{
    \includegraphics[width=0.32\textwidth]{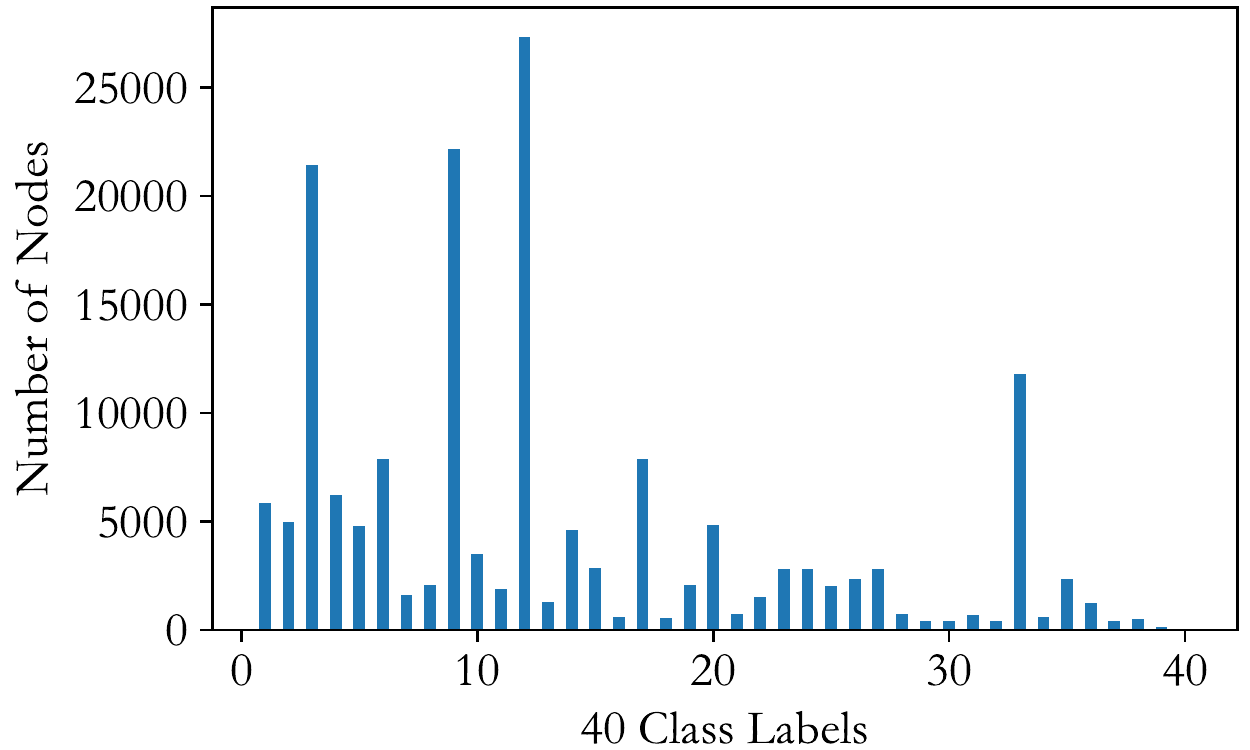}}
\subfigure[Class Distribution on arXivMath]{
    \includegraphics[width=0.32\textwidth]{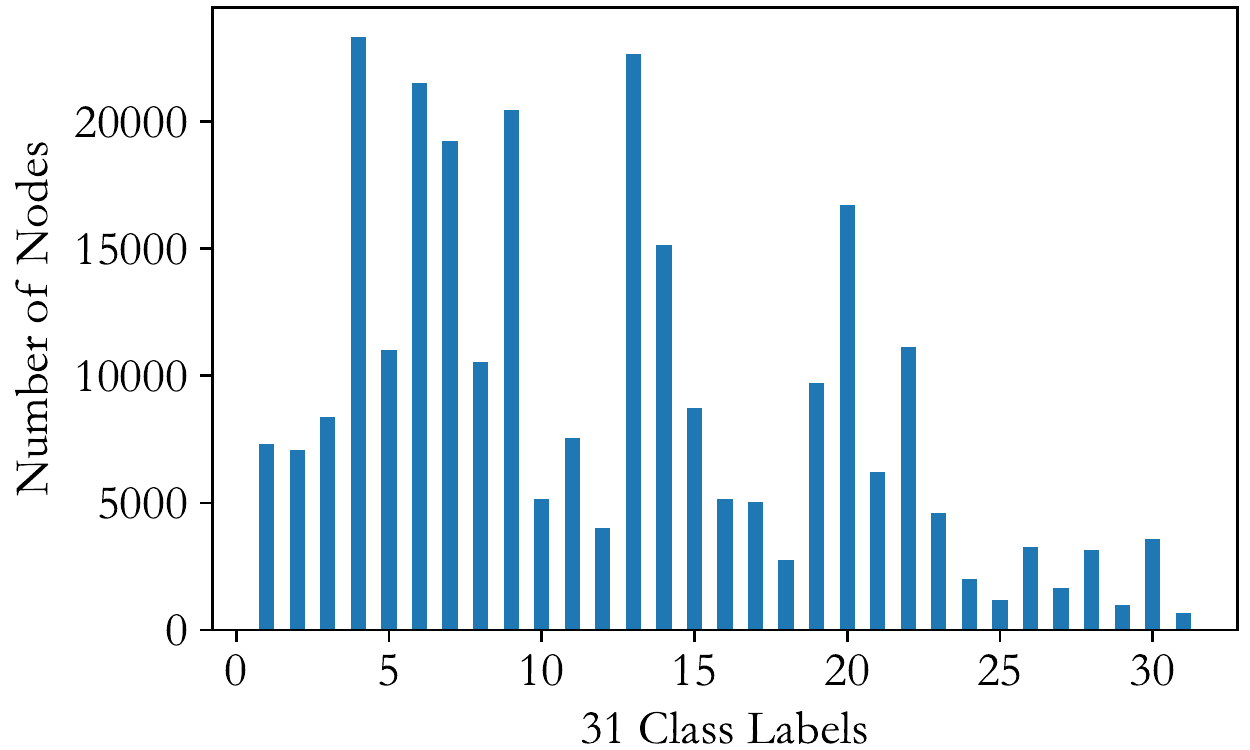}}
~\\
\subfigure[Class Distribution on arXivPhy]{
    \includegraphics[width=0.45\textwidth]{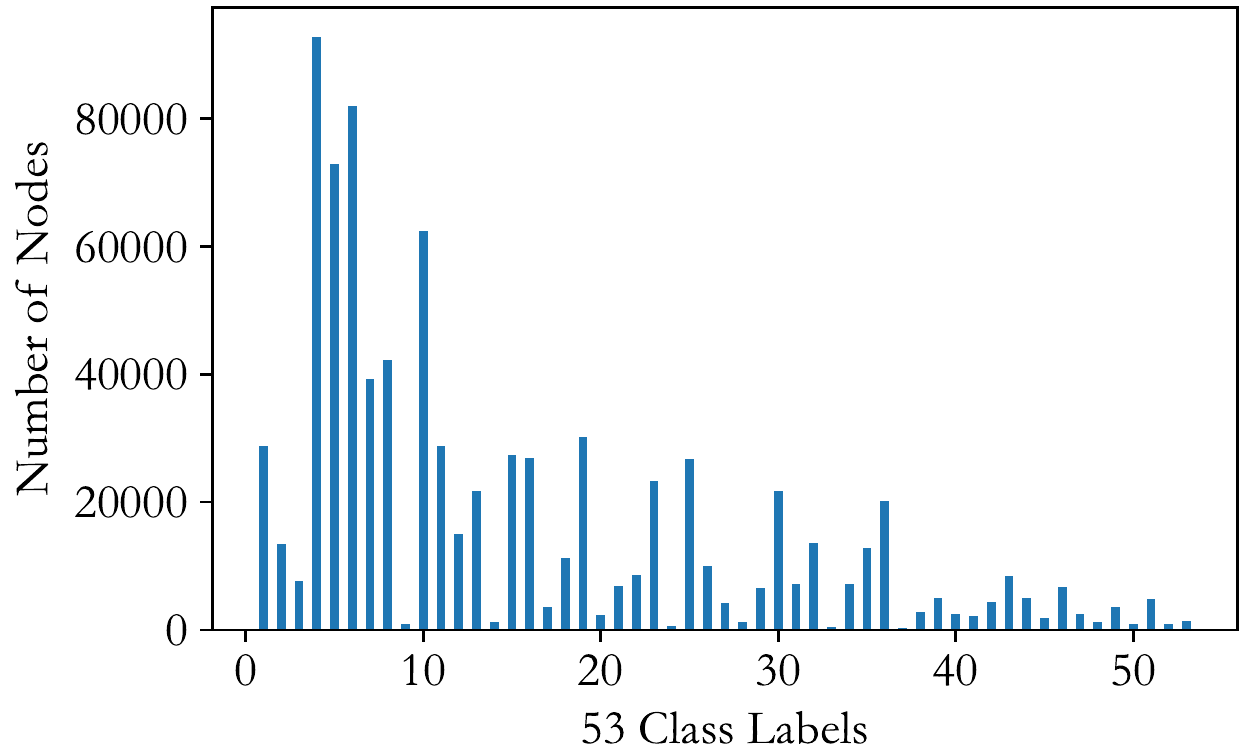}}
\subfigure[Class Distribution on arXivLarge]{
    \includegraphics[width=0.45\textwidth]{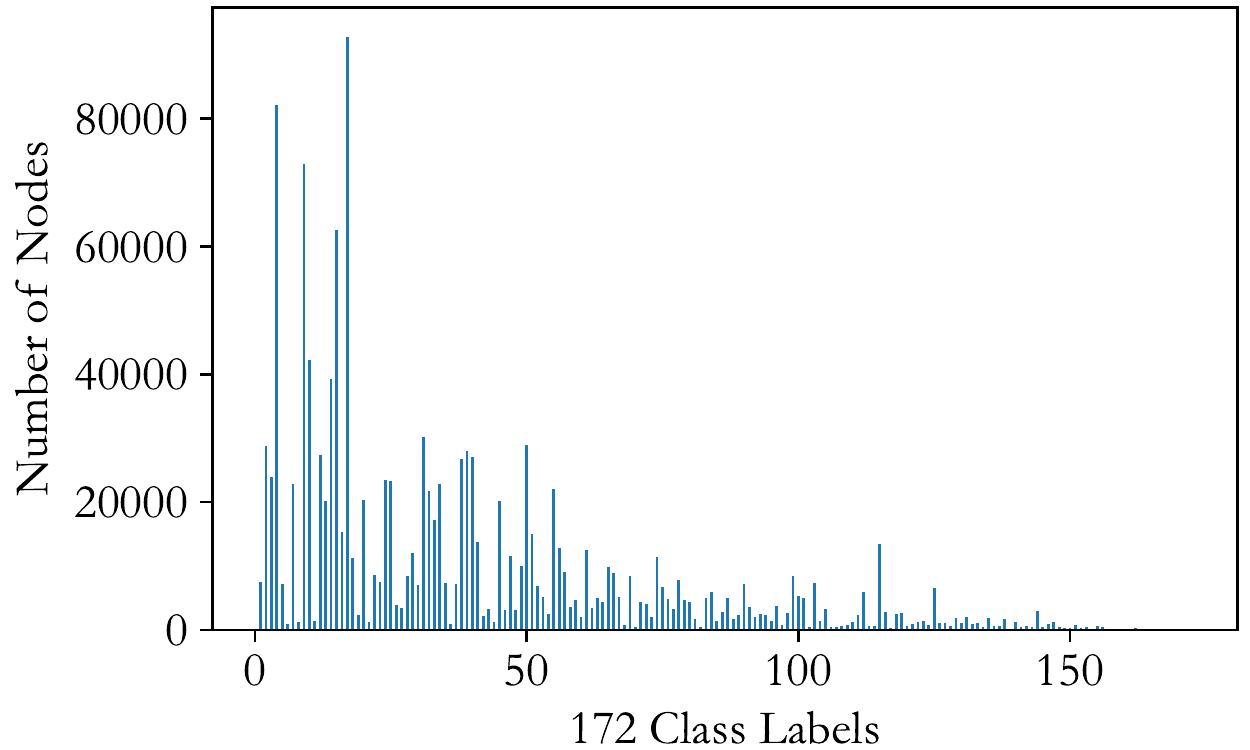}}
\caption{Class Distribution on arXiv4TGC datasets.}
\label{dis}
\end{figure}

Here we give the details of our five datasets. According to Table \ref{fields}, we report on different research subfields of arXiv4TGC datasets. Except for arXivLarge, which encompasses all 172 subfields listed by arXiv, the other four datasets are subsets of it, facilitating researchers to choose the data size more flexibly.

In addition, we also counted the distribution of classes for different graph datasets, as shown in Fig. \ref{dis}. We can find that the number of nodes corresponding to different classes varies widely. Because in real-world academic research, the number of researchers in different fields is inherently different. This data distribution is more in line with the real-world situation and can facilitate researchers to study the category imbalance problem.

Note that arXiv4TGC datasets are not only more suitable for temporal graph clustering but also can be used well for other tasks. And for these tasks, a set of reliably labeled large-scale temporal graph datasets can also be helpful for researchers.

\section{Experiments}
\label{Exp}

As mentioned above, we point out that arXiv4TGC datasets are not only suitable for temporal graph clustering but can also be naturally applied to other tasks. To verify it, we conduct experiments on temporal graph clustering and node classification, respectively.

\subsection{Baselines}

In this paper, we select several classical temporal graph learning models as baselines, including HTNE, TGC, TGN, TREND, and node2vec:







\textbf{HTNE} \cite{zuo2018embedding} is an early work for temporal graph learning, which utilizes the Hawkes process to model the neighborhood influence.

\textbf{TGC} \cite{TGC_ML} is an early framework for temporal graph clustering, we implement it here based on HTNE.

\textbf{TGN} \cite{rossi2020temporal} enables information updates during dynamic changes by storing information from historical interactions.

\textbf{TREND} \cite{wen2022trend} design the dynamic graph convolution network to aggregate high-order neighborhood information in temporal graphs.

\textbf{node2vec} \cite{grover2016node2vec} is a classical static graph method, which has been used as the pre-train model in TGC here. The node embeddings of node2vec have been generated during the TGC training process, so we will also evaluate them.

\subsection{Metrics and settings}
\label{settings}

For different tasks, we utilize different metrics. Temporal graph clustering is the main task in our paper, and we aim to demonstrate the superiority of our arXiv4TGC datasets by comparing the clustering performance of the baselines on different datasets. Here we utilize two types of evaluation metrics: external metrics and internal metrics.

Although temporal graph clustering works under unsupervised scenarios, when we want to evaluate the clustering performance, node labels are needed. External metrics measure clustering performance by calculating the difference between predicted and true labels, including NMI (normalized mutual information) and ARI (average rand index). NMI is a measure of how much information two sets of data share when clustered, and ARI is a measure of how similar two sets of clustered data are to each other. Both NMI and ARI have values between 0 and 1, with higher values indicating better clustering results. We argue that if multiple methods show poor results on the same dataset, then the dataset is not suitable for node clustering. 

In addition, there are also internal metrics that measure the clustering performance of datasets without node labels. These metrics mainly include CHI (calinski-harabasz index), and DBI (davies-bouldin Index). They are not as precise as external metrics, but they can be used as an auxiliary result reference. CHI measures the ratio of inter-cluster variance to intra-cluster variance, and DBI measures the inter-cluster distance and the intra-cluster distance, and it calculates the ratio of the sum of these distances to the number of clusters.

As an unsupervised task, the dataset used for node clustering does not need to be divided by training and test sets. We feed the whole dataset into temporal graph learning methods to generate node embeddings, which are then fed into the K-means model for evaluation.

For node classification, we construct a classifier to predict the node labels. The performance of the classifier is evaluated using metrics such as Accuracy and Weighted-F1. Accuracy measures the proportion of correctly classified nodes, while Weighted-F1 takes into account both the precision and recall of the classifier, weighted by the number of nodes in each class.

All the models are trained on the machine with Intel(R) i7-7700K CPU, 48G Memory, and NVIDIA RTX3090 GPU.

\subsection{Temporal graph clustering}

\begin{table}[t]
\caption{Node clustering performance on different datasets with external metrics. For the performance of each method on multiple datasets, we flag the top-3 results in red. Note that the training duration of TGN on arXivLarge and arXivPhy takes more than one month in RTX 3090Ti (24 GB), which we denote as TLE (time limit exceeded).}
\centering
\resizebox{\linewidth}{!}{
\begin{tabular}{c|cc|cc|cc|cc|cc}
    \toprule[2pt]
    Method     & \multicolumn{2}{c|}{HTNE} & \multicolumn{2}{c|}{TGC} & \multicolumn{2}{c|}{TGN} & \multicolumn{2}{c}{TREND} & \multicolumn{2}{c}{node2vec} \\
    Metric     & NMI         & ARI        & NMI        & ARI        & NMI        & ARI        & NMI         & ARI   & NMI         & ARI      \\
    \midrule[1pt]
    Bitcoin    & 8.34        & 1.44       & 8.91       & 1.99       & 6.44       & 1.24       & 6.84        & 0.46      & 8.41          & 2.31         \\
    ML1M       & 8.70        & 5.30       & 7.43       & 4.66       & 7.49       & 6.64       & 3.30        & 3.08      & 9.01          & 5.74        \\
    Amazon     & 0.36        & 1.23       & 0.49       & 0.63       & 0.13       & 0.78       & 0.10        & 0.18      & 0.25          & 0.03        \\
    Yelp       & 0.18        & 0.17       & 0.38       & 0.84       & 0.24       & 0.15       & 0.39        & 0.18      & 0.31          & 0.85        \\
    Tmall      & 10.54       & 5.54       & 10.72          & 5.96          & 6.63          & 4.72          & 5.07        & 3.28      & 10.57          & 7.97        \\
    \midrule[1pt]
    Brain      & \textcolor{red}{50.33}       & \textcolor{red}{29.26}      & \textcolor{red}{50.68}      & 30.03      & \textcolor{red}{41.71}      & \textcolor{red}{22.67}      & \textcolor{red}{45.64}       & 22.82      & \textcolor{red}{49.09}          & \textcolor{red}{28.40}       \\
    Patent     & 20.77       & 10.69      & 25.04      & 18.81      & 8.24       & 6.01       & 14.44       & 13.45      & 22.71          & 10.32       \\
    DBLP       & 35.95       & \textcolor{red}{22.13}      & 37.08      & 22.86      & \textcolor{red}{35.57}      & \textcolor{red}{19.40}      & 35.44       & 20.22      & 22.03          & 13.73       \\
    \midrule[1pt]
    arXivAI    & 39.24       & \textcolor{red}{43.73}      & 42.46      & \textcolor{red}{48.98}      & \textcolor{red}{24.74}      & 11.91      & 19.82       & \textcolor{red}{25.37}      & 34.34          & \textcolor{red}{36.08}       \\
    arXivCS    & 40.83       & 16.51      & 43.89      & \textcolor{red}{36.06}      & 16.21      & \textcolor{red}{18.63}      & 25.58       & \textcolor{red}{23.48}      & 40.86          & 14.03      \\
    arXivMath  & 36.81       & 12.77      & 42.54         & \textcolor{red}{25.69}          & 23.82      & 10.64      & 30.69   & 16.03      & 39.05          & 19.22        \\
    arXivPhy   & \textcolor{red}{50.25}       & 20.67      & \textcolor{red}{51.64}          & 24.29          & TLE          & TLE          & \textcolor{red}{41.27}       & \textcolor{red}{23.68}      & \textcolor{red}{50.33}          & \textcolor{red}{19.90}       \\
    arXivLarge & \textcolor{red}{50.79}       & 13.28      & \textcolor{red}{53.57}          & 21.91          & TLE          & TLE          & \textcolor{red}{30.75}       & 10.43      & \textcolor{red}{51.32}          & 13.52         \\
    \toprule[2pt]
\end{tabular}}
\label{performance}
\end{table}

As shown in Table \ref{performance}, we report the results on three types of datasets, i.e., labeled untrustworthy datasets, small-scale labeled datasets, and arXiv4TGC datasets (the unlabeled datasets cannot be used for clustering). We can find low performance obtained for the labeled untrustworthy datasets on the two main metrics of clustering performance, NMI and ARI, which implies that they are inherently unsuitable for clustering.

Note that we add the temporal graph clustering framework TGC \cite{TGC_ML} to HTNE (denoted as TGC in Table \ref{performance}), and the magnitude of the improvement of HTNE performance by TGC on the small-scale and arXiv4TGC datasets is more obvious than that on labeled untrustworthy datasets. These results show that although TGC can effectively improve the clustering results for temporal methods, the performance improvements on these labeled untrustworthy datasets are not significant enough. In other words, it means that if a method focusing on temporal graph clustering is tested on these datasets, even if performance gains are obtained compared to other methods, they may not be evident in the experimental results.

\begin{table}[t]
\caption{Node clustering performance on different datasets with internal metrics. For the performance of each method on multiple datasets, we flag the top-2 results in red. Note that we have marked the different evaluation criteria for the two metrics, the larger the CHI the better ($\nearrow$), and the smaller the DBI the better ($\searrow$).}
\centering
\begin{tabular}{c|cc|cc|cc}
    \toprule[2pt]
    Method     & \multicolumn{2}{c}{HTNE}         & \multicolumn{2}{c}{TREND} & \multicolumn{2}{c}{node2vec} \\
    Metric     & CHI ($\nearrow$)            & DBI ($\searrow$)            & CHI ($\nearrow$)          & DBI ($\searrow$)        & CHI ($\nearrow$)          & DBI  ($\searrow$)         \\
    \toprule[1pt]
    Bitcoin    & 497            & 1.7274          & 161            & 3.0151           & 129          & 2.9401        \\
    ML1M       & 1065           & 2.1237          & 13856          & \textcolor{red}{0.9541} & 994          & 2.1846        \\
    Amazon     & 3614           & 3.7426          & 5826           & 2.5196          & 2050         & 4.5547        \\
    Yelp       & 4241           & \textcolor{red}{1.4330}          & 2133           & 1.9700          & \textcolor{red}{36125}             &  2.1758             \\
    Tmall      & 4128           & 2.8930          & 3408           & 2.6954          & 6164             & 5.2099              \\
    \toprule[1pt]
    Brain      & 1671           & \textcolor{red}{1.0735} & 1985           & 1.6283          & 4091         & \textcolor{red}{1.1535}        \\
    Patent     & 1735           & 1.4546 & 2348           & 1.4432          & 1396         & \textcolor{red}{2.0209}        \\
    DBLP       & 1728           & 2.3878          & 760            & 3.167           & 555          & 3.7276        \\
    \toprule[1pt]
    arXivAI    & \textcolor{red}{11368} & 1.9304          & \textcolor{red}{22372} & \textcolor{red}{1.0718} & 3868         & 2.8461        \\
    arXivCS    & 7996           & 1.9530          & 13109          & 1.7665          & 2417         & 2.9301        \\
    arXivMath  & 5316           & 2.8577          & 5900           & 3.1534          & 3183         & 3.0088        \\
    arXivPhy   & \textcolor{red}{57767} & \textcolor{red}{1.2800} & \textcolor{red}{48305} & 2.0659          & \textcolor{red}{14405}        & 2.4121        \\
    arXivLarge & \textcolor{red}{42116} & 1.6264          & \textcolor{red}{32761} & \textcolor{red}{1.2303} &  \textcolor{red}{9803}            & \textcolor{red}{1.9264}        \\
    \toprule[2pt]   
\end{tabular}
\label{internal}
\vspace{-0.4 cm}
\end{table}

In addition, according to Table \ref{internal}, we can also find that in the internal metrics, our arXiv4TGC datasets can achieve better results overall in these datasets. It means that, for different methods, our datasets can provide a reliable potential clustering space, allowing these methods to obtain better clustering results.

Although the results are good on small-scale labeled datasets, they tend to be small to fit the large real-world data sizes and are therefore inappropriate for large-scale temporal graph clustering. In this case, our proposed arXiv4TGC dataset can show both good clustering performance and is suitable for large-scale clustering.

\subsection{Node classification}

\begin{figure}[t]
\centering
\subfigure[Node classification performance of HTNE]{
    \includegraphics[width=0.75\textwidth]{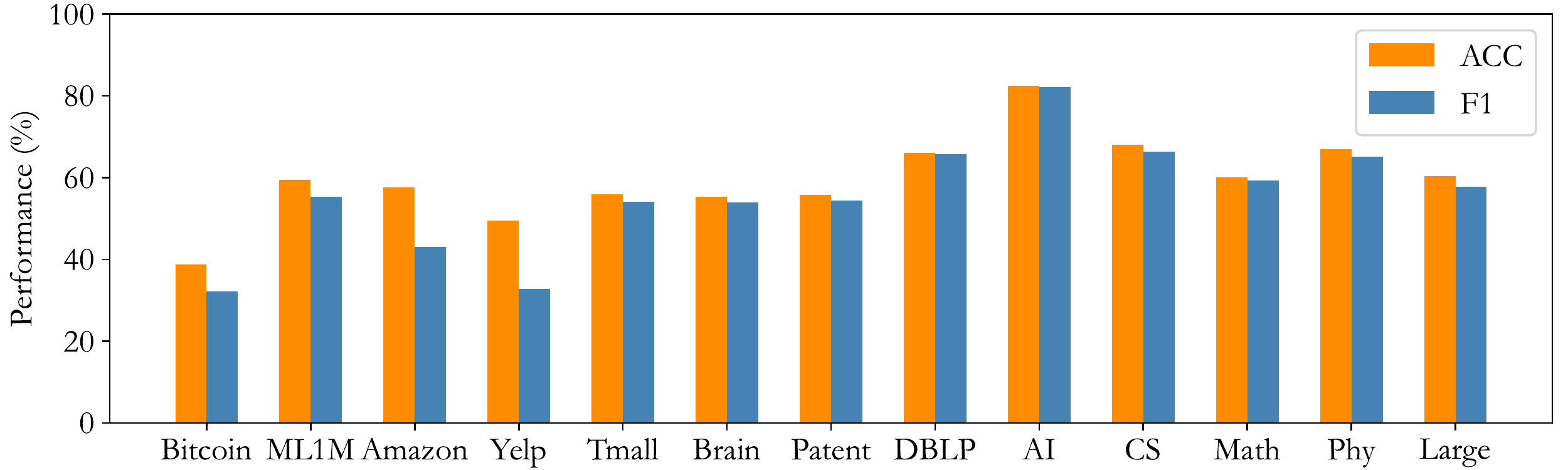}}
\subfigure[Node classification performance of TREND]{
    \includegraphics[width=0.75\textwidth]{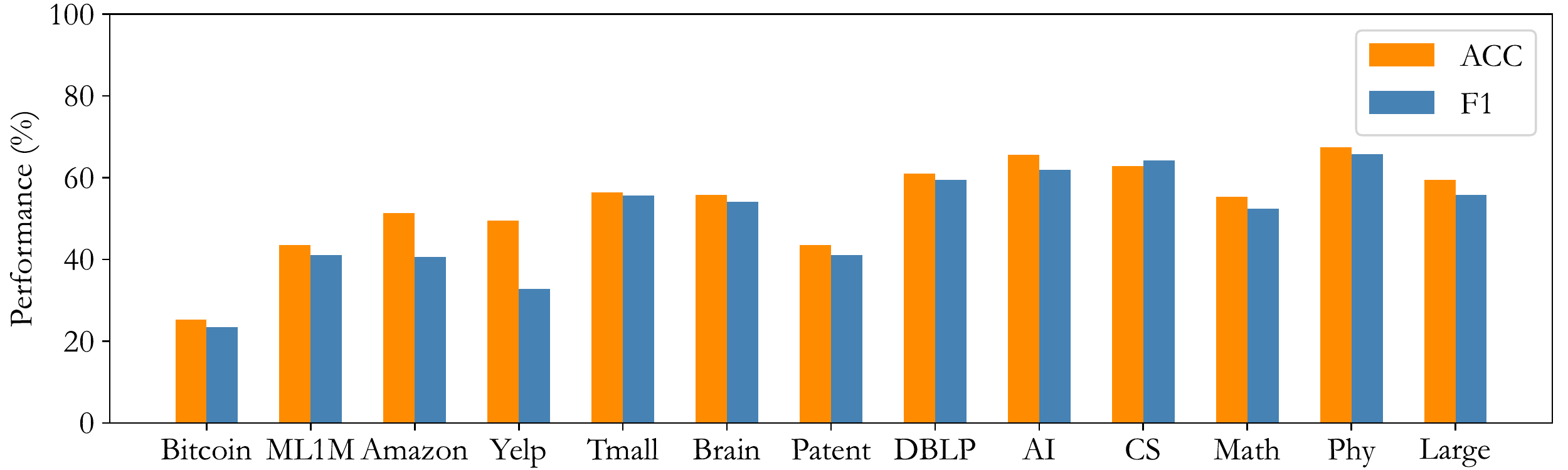}}
\subfigure[Node classification performance of node2vec]{
    \includegraphics[width=0.75\textwidth]{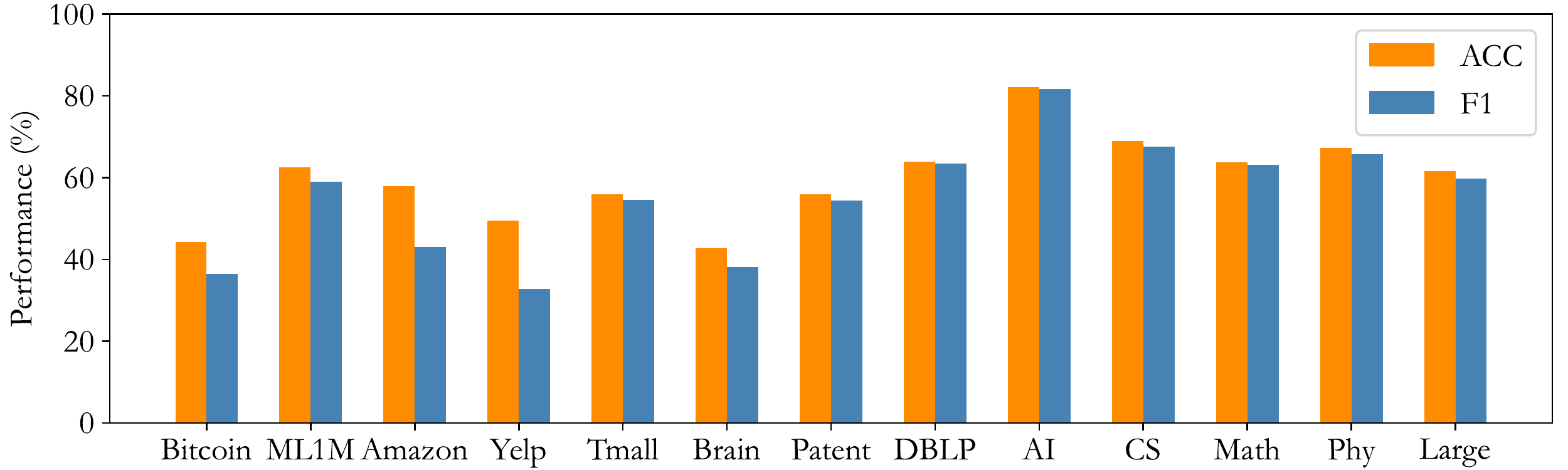}}
\caption{Node classification performance of several methods on different datasets. For presentation purposes, we have omitted the prefix 'arXiv' from arXiv4TGC datasets.}
\label{node class}
\end{figure}

As mentioned above, our arXiv4TGC datasets are not only suitable for temporal graph clustering but also for other graph tasks. In this section, we conducted a node classification task to verify this claim.

As shown in Fig. \ref{node class}, we report the node classification performance on different datasets. Our datasets tend to achieve better node classification results compared to other methods, which can be attributed to the presence of more reliable labels in our datasets.

Moreover, it is important to note that even datasets with unreliable labels can still achieve better classification results than clustering results. This is because the task settings for classification and clustering are different. Node classification requires the embedding of a specified category for each node, which is equivalent to selecting from multiple categories. Clustering, on the other hand, is more flexible and reflects the actual distribution of node features. Therefore, labels that are not credible in clustering still have a credible basis in the classification task.

In conclusion, our experiments on node classification demonstrate that our dataset is also applicable to other graph learning tasks. Due to space and training time constraints, we did not provide a performance comparison for link prediction. However, compared to node clustering and classification, link prediction focuses more on edge construction rather than node information. For academic graph datasets drawn from the real world data, their interactions are genuine and valid, which ensures the reliability of link prediction results, as exemplified by the effects of two academic datasets, DBLP and Patent. As our focus is on node division, we did not delve deeply into the results of link prediction.

\subsection{Limitation discussion}
\label{limit}

Indeed, there are certain limitations to our datasets. Firstly, their sizes could be expanded, but are currently limited by a lack of reliable node labels, which necessitates further integration of public resources. Secondly, the arXiv4TGC datasets also exhibit the class imbalance problem that is commonly observed in temporal graph datasets, which may adversely affect certain specialized tasks. Moreover, the information contained in these academic datasets is not rich enough to be easily applied to practical scenarios related to graph learning, such as fake news detection \cite{jin2022towards, ma2022curriculum}, knowledge graph \cite{SymCL, meng2023sarf}, social network analysis \cite{wasserman1994social, fan2022dynamic}, community detection \cite{jin2023predicting, fortunato2010community}, recommendation system \cite{jin2022code, wu2022adversarial}, bioinformatics application \cite{zang2023hierarchical, ABSLearn}, etc. Going forward, we intend to continue addressing these issues.

\section{Conclusion}

In this paper, we present a set of academic datasets for large-scale temporal graph clustering, denoted as arXiv4TGC. We observe that the absence of dependable node labels for large-scale temporal graph datasets impedes the advancement of temporal graph clustering. To explain this, we compare the scales of several publicly accessible datasets and the results of temporal graph clustering. Additionally, we delineate the processing and construction details of the arXiv4TGC datasets. Finally, we make these datasets and an implemented benchmark model available to the public.

Going forward, our focus will be on expediting the pace of temporal graph clustering training on large-scale datasets. Using K-means model to perform two-step clustering (first generating node embedding, then clustering) on large scale graph data requires a long evaluation time \cite{liuyue_Dink_net, wang2022highly}. Therefore, it is one of the future study directions to accomplish large-scale temporal graph clustering in one step (directly obtaining clustering results through model training).

\bibliographystyle{IEEEtran}
\bibliography{sample-base}

\end{document}